\documentclass[journal]{IEEEtran}
\ifCLASSINFOpdf
\else
\fi

\usepackage{amssymb}
\usepackage{latexsym}

\usepackage{url}
\usepackage{xcolor}
\usepackage{arydshln}
\usepackage{algorithm2e}
\usepackage{subcaption}
\usepackage{graphicx}
\usepackage{subcaption}
\usepackage{amsmath,amsfonts,amsthm,amssymb,bm}
\usepackage{t1enc}

\usepackage{amsmath,amssymb} 

\definecolor{newcolor}{rgb}{.8,.349,.1}

\begin{document}
%
\title{Filtering Point Targets via Online Learning of Motion Models}
%
%
%

\author{Mehryar~Emambakhsh,~Alessandro~Bay~and~Eduard~Vazquez
\thanks{M.~Emambakhsh and A.~Bay are with Cortexica Vision Systems Ltd., London, UK Email: \{mehryar.emambakhsh,~alessandro.bay\}@cortexica.com}
\thanks{E.~Vazquez is with AnyVision, Belfast, UK Email:  eduardov@anyvision.co}}

\maketitle

\begin{abstract}
Filtering point targets in highly cluttered and noisy data frames can be very challenging, especially for complex target motions. Fixed motion models can fail to provide accurate predictions, while learning based algorithm can be difficult to design (due to the variable number of targets), slow to train and dependent on separate train/test steps. To address these issues, this paper proposes a multi-target filtering algorithm which learns the motion models, on the fly, using a recurrent neural network with a long short-term memory architecture, as a regression block. The target state predictions are then corrected using a novel data association algorithm, with a low computational complexity. The proposed algorithm is evaluated over synthetic and real point target filtering scenarios, demonstrating a remarkable performance over highly cluttered data sequences.
\end{abstract}

\begin{IEEEkeywords}
Multi-target filtering and tracking, random finite sets, convolutional recurrent neural networks, long-short term memory, spatio-temporal data\end{IEEEkeywords}

%
\IEEEpeerreviewmaketitle

\section{Introduction}
Multi-target filtering (MTF) is the process of obtaining true positive samples from a cluttered and noisy data sequence. It has numerous applications in tracking \cite{Li:2018,Punchihewa:2018,Roy:2016}, radar/LiDAR signal processing \cite{Kulikov:2016}, simultaneous localization and mapping (SLAM) and occupancy grid computation in robotics, and sensor fusion \cite{Evers:2018,Leung:2017,Fantacci:2018,Li:20182,Xing:2016,Yan:2018}.\\
\indent
Defining a robust motion model is a key step for MTF algorithms \cite{Roth:2014,Li:2000,Zhai:2014}. Briefly, motion models formulate the prior knowledge about the variations over the state (latent) space. In a Bayesian framework, they are used to predict the target states which are then corrected using the obtained measurements (observations). A weak motion model can deteriorate the filtering performance by propagating a wrong prediction over the state space. Such issue can be even more salient for fixed motion models, which do not adapt themselves to the incoming data.
On the other hand, learning such motion patterns can be difficult, because: (1) In an MTF problem, the number of targets are usually variable and unknown, making the model design very difficult; (2) Since the incoming data sequence is usually highly cluttered and noisy, learning-based models can be trained on false positive samples creating mis-information propagation; (3) Due to the high number of parameters influencing filtering problems, assigning separate train/validation and test scenarios is very difficult and can lead the model to over-fit. The motion model is expected to be learned online, using the incoming data until the current time step; (4) Speed is crucial; a learning-based method should be computationally comparable with its fixed motion model-based rivals.

Considering these challenges, in this paper we propose on-line learning of the motion models (OLMM) to perform MTF from the incoming sequence of data. This is performed, on the fly, by training recurrent neural networks (RNN) with long short-term memory (LSTM) architecture, used as a regression block, over the target state space. 
The filtering and update is then performed by a novel data association algorithm.
Our implementation allows GPU memory re-usability by {\emph{placeholders}} utilisation and facilitates transfer learning by initialising the LSTM state predictors by reusing weights and biases from other targets.
We have evaluated the algorithm over two datasets containing point targets: (1) A commonly used synthetic data, which contains numerous MTF challenges, such as non-linear motion, birth, spawn, merge and death of targets; (2) The bird's-eye view of the Duke Multi-Target, Multi-Camera (DukeMTMC) pedestrian tracking dataset.
Our experimental results show a remarkable performance of our algorithm when compared with previous filtering approaches.

\textbf{\textit{Contributions}.}
Unlike the MTF algorithms in ~\cite{Reuter:2014,Vo:2014}, which use fixed motion models, the proposed algorithm learns the motion model from the incoming data. The proposed data association algorithm has linear complexity and compared with RFS MTF algorithms \cite{Reuter:2014,Vo:2014}, relies on significantly fewer number of hyper parameters.
For example, \cite{Vo:2006} requires hyper parameters to perform pruning, merge and truncation of the output density function, in addition to clutter distribution, survival and detection probabilities. As opposed to the previous neural network-based methods \cite{Milan:2017}, OLMM does not rely on a separate training and test steps and is trained on the fly.
To the best of our knowledge, the proposed algorithm is one of the first of its kind, which is capable of applying LSTM to filter a densely clutter sequence of data. It should be mentioned that the current paper is an extension of our recent work in \cite{Emambakhsh:2019}, which has been significantly enhanced in the following aspects: (1) More extensive experimental results are provided over both synthetic and real data; (2) Compared to our initial paper, a significantly improved and unified mathematical framework is provided. Algorithm is explained via several diagrams and a pseudo code for immediate implementation is provided; (3) Complexity analysis and elapsed time for each time step are reported.

In this paper, we use italic font for scalar, tuple and random finite set (RFS) parameters/variables, while bold font is used for vectors and matrices. Also, we use $t$, $k$ and $k^\prime$ to indicate the time step, sample index from target and measurement RFS, respectively.

\section{Related work}
\label{sec:relWork}
\subsection{Fixed models: Bayesian paradigms}
Prior modelling of the targets' behaviour can be based on appearance or motion (kinematics) equations. Using these models the state vectors defined for each target are predicted. Then the predictions are mapped onto the measurement step to perform correction. 
In a Bayesian formulation of a single-target filtering, the goal is to estimate the (hidden) target state, from a set of observations.
Filtering is a recursive problem; The state estimation at the $t^{th}$ time step is usually obtained by Maximum A Posteriori (MAP) criterion, over the state space given the past observations.
Kalman filter is arguably the most popular online filtering approach. It assumes linear motion models with Gaussian distributions for both the prediction and update steps. Non-linearity and non-Gaussian behaviour are addressed by Extended and Unscented Kalman Filters {(EKF, UKF)}, respectively. 

Mahler proposed RFS \cite{Mahler:2003}, an encapsulated formulation for MTF, incorporating clutter density, probabilities of detection, survival and birth of targets \cite{Vo:2005,Mahler:2003,Mahler:2007}. Targets and measurements are assumed to form sets, with variable random cardinalities. Using Finite Set Statistics \cite{Mahler:2003}, the posterior distribution for a single target can be extended from vectors to RFS.
Facilitated by the RFS formulation, Probability Hypothesis Density (PHD) maps \cite{Vo:2005,Vo:2006} are proposed to represent target states. These maps have two basic features: 1) Their peaks correspond to the location of targets; 2) Their integration gives the expected number of targets at each time step. In their seminal paper, Vo and Ma proposed Gaussian Mixture PHD (GM-PHD), which propagates the first-order statistical moments to estimate the posterior density as a mixture of Gaussians \cite{Vo:2006}.

While GM-PHD represents the hypothetical target state via mixture of Gaussians, a particle filter-based solution is proposed by Sequential Monte Carlo PHD (SMC-PHD) to address non-Gaussian distributions \cite{Vo:2005}. 
Cardinalised PHD (CPHD) is proposed by Mahler to also propagate the cardinality of the targets over time in \cite{Mahler:2007}, while its intractability is addressed in \cite{Nagappa:2017}. Also, Lu {\emph{et al.}} proposed an algorithm addressing missed detections, enhancing the track continuity \cite{Lu:2017}. On the other hand, target spawning within CPHD framework is addressed in \cite{Bryant:2017}.
A PHD and CPHD filter which propagates the second order statistics in parallel with the mean is proposed by Schlangen \emph{et al.} \cite{Schlangen:2018}, which significantly outperforms CPHD in terms of computational cost.

The Labelled Multi-Bernoulli Filter (LMB) is introduced in \cite{Reuter:2014}, which performs track-to-track association and outperforms previous algorithms in the sense of not relying on high signal to noise ratio. 
Vo \emph{et al.} proposed Generalized Labelled Multi-Bernoulli (GLMB) as a labelled MTF \cite{Vo:2014}, while Garc\'{i}a-Fern\'{a}ndez \emph{et al.} introduced an approach to derive Poisson LMB without using the probability generating functionals \cite{Garc:2018}.
Since a large number of particles needs to be propagated during Monte Carlo based methods, the computational complexity can be high and hence gating might be necessary. An inaccurate gating, however, can filter out legitimate targets and increase the false negative rate.

\subsection{Neural filtering}
Parisini and Zoppoli reformulated the process of state estimation of MTF algorithms to a non-linear programming problem~\cite{Parisini:1994}. Although since then, the neural network based sequential learning solutions were infamous for their vulnerability to small datasets, easily under-/over-fitting and slow computational speed during test and train phases, with the advances in computation power in recent years, neural network-based approaches have been capable of learning from large number of sequences. This has opened a new window for MTF as these methods are capable of modelling the latent information within the data sequence in parallel with filtering its false positive samples.
RNNs are neural networks with feedback loops, through which past information can be stored and exploited. They offer promising solutions to difficult tasks such as system identification, prediction, pattern classification, and stochastic sequence modelling \cite{Bay2016}.
RNNs are known to be particularly hard to train, especially when long temporal dependencies are involved, due to the so-called vanishing gradient phenomenon.
Learning motion models via neural filters can be difficult, because of the varying number of targets in a cluttered scene, which is quite common in an MTF problem. This can make the model design very difficult, especially since neural networks usually have a fixed architecture. Also, since the incoming data sequence is usually highly cluttered and noisy, learning-based models can be trained on false positive samples creating mis-information propagation. 
Assigning separate train/validation and test scenarios is very difficult for filtering scenarios. This makes neural filtering algorithms vulnerable to over-fitting. And finally, the motion model is expected to be learned online, using the available data at the current time step, which is another challenge for neural filtering. 

\begin{figure*}[!t]
\centering
\includegraphics[width=0.8\textwidth]{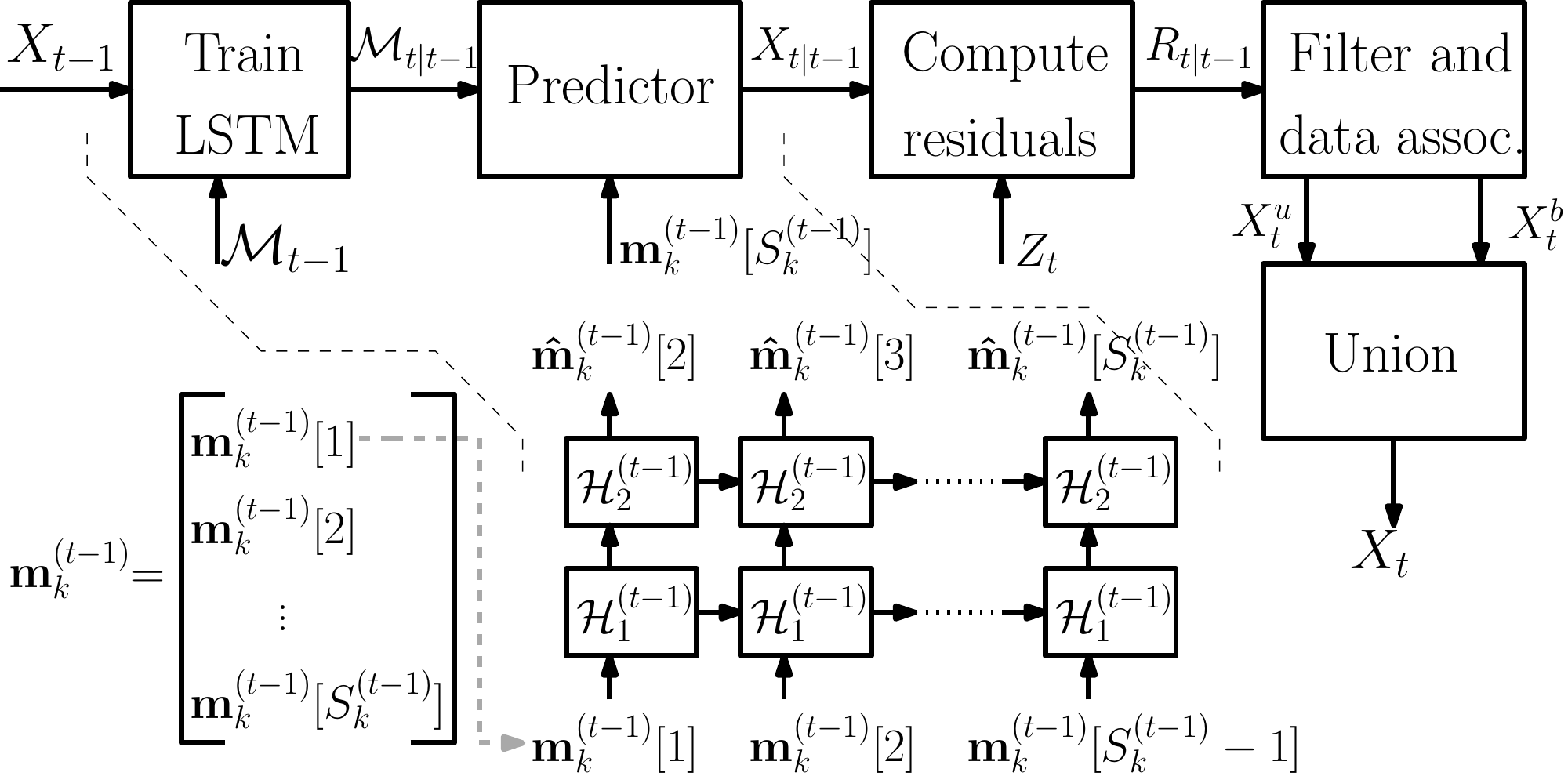}
\caption{The overall MTF pipeline: The LSTM network is trained using the previous target batch, which then predicts the target state for the current time step. The prediction is then updated and filtered using the obtained measurement set, in which survival, death and birth of targets are assigned.}
\label{fig:overall}
\end{figure*}

\section*{List of symbols}
\begin{itemize}
\item[] $t$: time step
\item[] $k$: target index
\item[] $k^{\prime}$: measurement (observation) index
\item[] $x^{(t)}_k$: $k^{th}$ target tuple at time $t$
\item[] $X_t$: target RFS at $t$
\item[] $M_t$: number of targets at $t$
\item[] ${\bf{m}}^{(t)}_{k}$: $k^{th}$ target state matrix at $t$
\item[] $S^{(t)}_k$: number of collected sample for the $k^{th}$ target at $t$
\item[] $d$: dimensionality of the state space
\item[] $a^{(t)}_k$: age of the $k^{th}$ target at $t$
\item[] $g^{(t)}_k$: {\emph{genuinity}} error of the $k^{th}$ target at $t$
\item[] $f^{(t)}_k$: {\emph{freeze}} state of the $k^{th}$ target at $t$
\item[] $\mathcal{H}^{(t)}_l$: $l^{th}$ layer of LSTM model at $t$
\item[] $L$: number of LSTM network hidden layers
\item[] ${\bf{h}}^{(t)}_{l}$: hidden state of the $l^{th}$ layer of LSTM model at $t$
\item[] ${\bf{i}}^{(t)}_{l}$: input gate of the $l^{th}$ layer of LSTM model at $t$
\item[] ${\bf{j}}^{(t)}_{l}$: transform gate of the $l^{th}$ layer of LSTM model at $t$
\item[] ${\bf{f}}^{(t)}_{l}$: forget gate of the $l^{th}$ layer of LSTM model at $t$
\item[] ${\bf{o}}^{(t)}_{l}$: output gate of the $l^{th}$ layer of LSTM model at $t$
\item[] ${\bf{c}}^{(t)}_{l}$: memory cell of the $l^{th}$ layer of LSTM model at $t$
\item[] $\phi^{(\bullet)}()$: element-wise activation function for LSTM gate~$\bullet$
\item[] $\mathcal{M}_t$: LSTM model tuple at $t$
\item[] $\mathcal L$: loss function
\item[] ${\bf{\hat m}}_k^{(t)}$: $k^{th}$ target's estimated state at $t$
\item[] $\tau$: target index within the state matrix
\item[] $\mathcal{M}_{t|t-1}$: number of predicted targets
\item[] $x^{(t|t-1)}_k$: $k^{th}$ target predicted tuple
\item[] $X_{t|t-1}$: predicted target RFS
\item[] $Z_t$: measurement RFS at $t$
\item[] $N_t$: number of measurements at $t$
\item[] $r^{(t|t-1)}_{k, k^\prime}$: residual tuple computed using the target $k$ and measurement $k^\prime$ at $t|t-1$
\item[] $R_{t|t-1}$: residual RFS at $t|t-1$
\item[] $T^{(t|t-1)}_{k^\prime, k}$: {\emph{targetness}} error calculated using the target $k$ and measurement $k^\prime$ at $t|t-1$
\item[] ${\bf{z}}^{(t)}_{k^\prime}$: ${k^\prime}^{th}$ measurement at $t$
\item[] ${\bm{\mathcal{T}}}^{(t|t-1)}$: targetness error matrix at $t|t-1$
\item[] $\mathcal{C}^I_{t|t-1}(k^\prime)$: index of the closest target to the ${k^\prime}^{th}$ measurement at $t|t-1$
\item[] $\mathcal{R}^I_{t|t-1}(k)$: index of the closest measurement to the $k^{th}$ target at $t|t-1$
\item[] $\mathcal{C}_{t|t-1}(k^\prime)$: distance of the closest target to the ${k^\prime}^{th}$ measurement at $t|t-1$
\item[] $\mathcal{R}_{t|t-1}(k)$: distance of the closest measurement to the $k^{th}$ target at $t|t-1$
\item[] $\mathcal{C}^I_{t|t-1}$: vector containing each $\mathcal{C}^I_{t|t-1}(k^\prime)$ for $k^\prime = \{1, 2, \ldots, N_t\}$
\item[] $\mathcal{R}^I_{t|t-1}$: vector containing each $\mathcal{R}^I_{t|t-1}(k)$ for $k = \{1, 2, \ldots, M_{t|t-1}\}$
\item[] $\mathcal{H}^{\mathcal{C}}_{t|t-1}$: histogram of $\mathcal{C}^I_{t|t-1}$
\item[] $\mathcal{H}^{\mathcal{R}}_{t|t-1}$: histogram of $\mathcal{R}^I_{t|t-1}$
\item[] $\lambda_c$: mean of the Poisson distribution
\item[] $\sigma_r$: standard deviation of the radial detection error 
\item[] $\sigma_\theta$: standard deviation of the bearing detection error
\item[] $a_{min}$: minimum target age
\item[] $g_{min}$: minimum target genuinity error
\item[] $g_{max}$: maximum target genuinity error
\item[] $X^u_t$: updated target RFS at $t$
\item[] $X^b_t$: birth target RFS at $t$
\end{itemize}

\section{Target representation}
\label{sec:targetRep}
Let us define $x_k^{(t)}$, the $k^{th}$ target tuple at the $t^{th}$ time step as a member of RFS $X_t = \left\{x_1^{(t)}, x_2^{(t)}, \ldots, x_k^{(t)}, \ldots, x_{M_t}^{(t)}\right\}$ as follows,
\begin{equation}
\begin{array}{l}
x_k^{(t)} = \\ \left({\bf{m}}_k^{(t)} \in \mathbb{R}^{S_{k}^{(t)} \times d}, a_k^{(t)} \in \mathbb{Z}^{1\times1}, g_k^{(t)} \in \mathbb{R^{+}}^{1\times1}, f_k^{(t)} \in \{0, 1\}\right),
\label{eq:xl}
\end{array}
\end{equation}
\noindent where ${\bf{m}}_k^{(t)}$ (an $S_{k}^{(t)} \times d$ matrix) contains target state of $S_{k}^{(t)}$ samples over a $d$-dimensional state space. $a_k^{(t)}$ is an integer indicating the age of the $k^{th}$ target (the higher the age, the longer the target has survived). $g_k^{(t)}$ is a real positive number containing the target's {\emph{genuinity}} error; It quantifies how legitimate the current target is over a continuous space, where its higher values correspond to higher likelihood of false positivity. $f_k^{(t)}$ is a binary {\emph{freeze}} state variable, which is: 1 (True), if there is no associated measurement for this target (due to occlusions, false positivity or detection failure); 
Or 0 (False), when there is at least one measurement associated with this target (a ``surviving'' target). $X_t$ is an RFS with $M_t$ cardinality, which contains all the target tuples at $t$.

\section{OLMM pipeline}
\label{sec:MTFPip}
The overall MTF pipeline in one time step is illustrated as a block diagram in Fig.~\ref{fig:overall}. The LSTM network is trained using the available data for each target and then used to predict target state. Next, a set of residuals is computed over the predictions and current measurement sets. Filtering and data association are finally performed to assign target survival and birth sets. In the following sections, each of these steps is explained in detail.

\subsection{Online motion modelling}\label{sec:LSTMTrain}
The target state variations over video frames can be seen as a sequential learning problem. Thus, we apply an LSTM network to learn a global motion model, since \emph{dedicating} one LSTM for each target leads to memory management issues. The network is trained online for each target using its past measurements, transferring the learned weights and biases from one to the other target. Formally, we used an $L$-layer LSTM defined as,
\begin{align}
&\mathcal{H}^{(t-1)}_l :\\
& \begin{cases}
\textbf{i}^{(t-1)}_l[\tau] = \phi^{(i)}\bigg(\textbf{A}_l^{(i)} \textbf{h}_{l-1}^{(t-1)}[\tau] + \textbf{B}_l^{(i)} \textbf{h}_l^{(t-1)}[\tau-1] +\textbf{b}_l^{(i)} \bigg) \\
\textbf{j}^{(t-1)}_l[\tau] = \phi^{(j)}\bigg(\textbf{A}_l^{(j)} \textbf{h}_{l-1}^{(t-1)}[\tau] + \textbf{B}_l^{(j)} \textbf{h}_l^{(t-1)}[\tau-1] +\textbf{b}_l^{(j)} \bigg) \\
\textbf{f}^{(t-1)}_l[\tau] = \phi^{(f)}\bigg(\textbf{A}_l^{(f)} \textbf{h}_{l-1}^{(t-1)}[\tau] + \textbf{B}_l^{(f)} \textbf{h}_l^{(t-1)}[\tau-1] +\textbf{b}_l^{(f)} \bigg) \\
\textbf{o}^{(t-1)}_l[\tau] = \phi^{(o)}\bigg(\textbf{A}_l^{(o)} \textbf{h}_{l-1}^{(t-1)}[\tau] + \textbf{B}_l^{(o)} \textbf{h}_l^{(t-1)}[\tau-1] +\textbf{b}_l^{(o)} \bigg) \\
\textbf{c}^{(t-1)}_l[\tau] = \textbf{c}^{(t-1)}_l[\tau-1] \odot \textbf{f}^{(t-1)}_l[\tau] + \textbf{i}^{(t-1)}_l[\tau] \odot \textbf{j}^{(t-1)}_l[\tau] \\
\textbf{h}^{(t-1)}_l[\tau] = \tanh(\textbf{o}_l^{(t-1)}[\tau]) \odot \textbf{c}_l^{(t-1)}[\tau]
\end{cases}\nonumber \\
&{\bf{\hat m}}_k^{(t-1)}[\tau+1] = \phi^{(y)} \bigg( \textbf{A}^{(y)} \textbf{h}_L^{(t-1)}[\tau] + \textbf{b}^{(y)} \bigg), \label{eq:LSTM_out}
\end{align}
where, for each layer $l=1,\ldots,L$, each hidden block $\mathcal{H}_l^{(t-1)}$ computes the hidden state $\textbf{h}_l^{(t-1)}$, using four gates $\textbf{i}_l^{(t-1)}$ , $\textbf{j}_l^{(t-1)}$, $\textbf{f}_l^{(t-1)}$, $\textbf{o}_l^{(t-1)}$ (i.e. input, transform, forget, and output gates, respectively). The non-linear element-wise activation functions are defined as $\phi^{(i)},\phi^{(j)},\phi^{(f)},\phi^{(o)}$, while $\textbf{c}_l$ is the memory cell. Then, the network estimates the next target state as ${\bf{\hat m}}_k^{(t-1)}[\tau+1]$, for $\tau=1,\ldots,S_k^{(t-1)}-1$, using the last hidden state and a linear element-wise activation function $\phi^{(y)}$. Therefore, the network is completely described by the model tuple $\mathcal{M}_t$, containing the weights and biases of the network: $\mathcal{M}_{t} = \left(\textbf{A}_l^{(\bullet)},\textbf{B}_l^{(\bullet)},\textbf{b}_l^{(\bullet)},\textbf{A}^{(y)},\textbf{b}^{(y)}\right)$ (for simplicity of notation, we have omitted $t$ for the weight matrix and bias vectors within the model tuple). Thus, given the model tuple at the previous time step $\mathcal{M}_{t-1}$, the network is updated as a regression block to minimise a mean square error loss function $\mathcal{L}$ as follows,
\begin{align}
&\mathcal L = \frac{1}{S_k^{(t-1)}-1}\nonumber \times\\ 
&\sum_{\tau = 2}^{S_k^{(t-1)}} \bigg({\bf{\hat m}}_k^{(t-1)}[\tau]-{\bf{m}}_k^{(t-1)}[\tau]\bigg)\bigg({\bf{\hat m}}_k^{(t-1)}[\tau]-{\bf{m}}_k^{(t-1)}[\tau]\bigg)^\intercal,
\end{align}
which is calculated over the new estimated target state ${\bf{\hat m}}_k^{(t-1)}[\tau]$ and the expected ${\bf{m}}_k^{(t-1)}[\tau]$, for $\tau = 2,\ldots,S_k^{(t-1)}$. Minimising $\mathcal{L}$ gives the new model parameters ($
\mathcal{M}_{t|t-1} = \underset{\mathcal{M}_{t-1}}{\mathrm{argmin}} \left( {\mathcal{L}} \right)
$).
$\mathcal{M}_{t|t-1}$ contains the updated weights and biases of LSTM.

\subsection{Predicting the target state}
After the LSTM network is trained, we use the updated weights $\mathcal{M}_{t|t-1}$ and the latest target state vector ${\bf{m}}_k^{(t-1)}[S_k^{(t-1)}]$ to compute the predicted target state ${\hat{\bf{m}}}_{k}^{(t|t-1)} = {\bf{\hat m}}_k^{(t-1)}[S_k^{(t-1)}+1]$. 
This procedure is repeated for all targets in $X_t$, resulting in the following predicted RFS,
\begin{equation}
\begin{array}{l}
X_{t|t-1} = \left\{x_{1}^{(t|t-1)}, \ldots, x_{k}^{(t|t-1)} \ldots, x_{M_{t|t-1}}^{(t|t-1)}\right\}\\
x_{k}^{(t|t-1)} = \left(x_{k}^{(t-1)}, {\hat{\bf{m}}}_{k}^{(t|t-1)} \right)
\end{array},
\end{equation}
where $x_{k}^{(t|t-1)}$ is the $k^{th}$ prediction tuple and $X_{t|t-1}$ is the predicted RFS.

\subsection{Filtering and update}
\noindent\textbf{\textit{Computing residuals.}}
At the $t^{th}$ time step, a set of residuals are calculated using the obtained measurement RFS $Z_t$. If $Z_t$ has $N_t$ cardinality, assuming no gating is performed, there will be $N_t \times M_{t|t-1}$ residuals which are stored as $R_{t|t-1} = \{r_{1, 1}^{(t|t-1)}, \ldots, r_{k, k^\prime}^{(t|t-1)}, \ldots, r_{M_{t|t-1}, N_t}^{(t|t-1)}\}$, where its ${k \times k^\prime}^{th}$ tuple $r_{k, k^\prime}^{(t|t-1)} \in R_{t|t-1}$ contains the residual information between the $k^{th}$ target and $k^\prime$ measurement as follows,
\begin{equation}
r_{k, k^\prime}^{(t|t-1)} = \left(x_k^{(t|t-1)}, T_{k^\prime, k}^{(t|t-1)} \in {\mathbb{R^{+}}}^{1 \times 1}, {\bf{z}}_{k^\prime}^{(t)} \in Z_t\right),
\label{eq:myres}
\end{equation}
\noindent in which $Z_t = \{{\bf{z}}_{1}^{(t)}, {\bf{z}}_{2}^{(t)}, \ldots, {\bf{z}}_{k^\prime}^{(t)}, \ldots, {\bf{z}}_{N_t}^{(t)}\}$ is the measurement RFS, ${\bf{z}}_{k^\prime}^{(t)}$ is the ${k^\prime}^{th}$ ($1 \times d$) measurement vector and $T_{k, k^\prime}^{(t|t-1)}$ is the \emph{targetness} error parameter, which is computed as the second norm between the measurement and target state as follows,
\begin{equation}
T_{k^\prime, k}^{(t|t-1)} = \left\lVert {\bf{z}}_{k^\prime}^{(t)} - {\hat{\bf{m}}}_{k}^{(t|t-1)} \right\rVert_{2}. 
\label{eq:residualEQ}
\end{equation}
$T_{k^\prime, k}^{(t|t-1)}$ is a distance metric between the predicted target vector ${\hat{\bf{m}}}_{k}^{(t|t-1)}$ and measurement vector ${\bf{z}}_{k^\prime}^{(t)}$. 

$R_{t|t-1}$ is used to perform the filtering step, at which survival of targets are determined, new births are assigned and false positive targets and measurements are removed.
To do this, first using (\ref{eq:residualEQ}), an $N_t \times M_{t|t-1}$ matrix ${\bm{\mathcal{T}}}^{(t|t-1)}$ is constructed as follows,
\[
{\bm{\mathcal{T}}}^{(t|t-1)} = \begin{bmatrix} 
    T_{1, 1}^{(t|t-1)} & \dots & T_{1, M_{t|t-1}}^{(t|t-1)} \\
    \vdots & \ddots & \vdots \\
    T_{N_t, 1}^{(t|t-1)} & \ldots & T_{N_t, M_{t|t-1}}^{(t|t-1)} 
    \end{bmatrix}
\]
whose element at the ${k^{\prime}}^{th}$ row and $k^{th}$ column gives $T_{k^\prime, k}^{(t|t-1)}$. ${\bm{\mathcal{T}}}^{(t|t-1)}$ contains the targetness errors between all measurements and target states.
In the next section, we detail how ${\bm{\mathcal{T}}}^{(t|t-1)}$ is used to perform data association.

\noindent\textbf{\textit{Data association.}}
For each column and row of ${\bm{\mathcal{T}}}^{(t|t-1)}$ the measurement and target indexes corresponding to the lowest $T_{k^\prime, k}^{(t|t-1)}$ are computed, respectively, as follows,
\begin{equation}
\begin{array}{l}
\mathcal{C}^{I}_{t|t-1} (k') = \underset{k}{\mathrm{argmin}} \left(T_{k^\prime, k}^{(t|t-1)}\right)\; , \; k^\prime = 1, \ldots, N_t \\
\mathcal{R}^{I}_{t|t-1} (k) = \underset{k^\prime}{\mathrm{argmin}} \left(T_{k^\prime, k}^{(t|t-1)}\right)\; , \; k = 1, \ldots, M_{t|t-1}
\end{array}
\end{equation}
\noindent where $\mathcal{C}^{I}_{t|t-1} = \left[\mathcal{C}^{I}_{t|t-1} (1), \mathcal{C}^{I}_{t|t-1} (2), \ldots, \mathcal{C}^{I}_{t|t-1} (N_t)\right]^\intercal$ and \\ \noindent $\mathcal{R}^{I}_{t|t-1} = \left[\mathcal{R}^{I}_{t|t-1} (1), \mathcal{R}^{I}_{t|t-1} (2), \ldots, \mathcal{R}^{I}_{t|t-1} (M_{t|t-1}) \right]$ are $N_k \times 1$ and $1 \times M_{k-1}$ vectors, containing the indexes of the closest target and measurement, respectively. In addition to their indexes, the corresponding minimum values of each row and column of ${\bm{\mathcal{T}}}^{(t|t-1)}$ are also computed,
\begin{equation}
\begin{array}{l}
\mathcal{C}_{t|t-1} (k') = \underset{k}{\mathrm{min}} \left(T_{k^\prime, k}^{(t|t-1)}\right)\; , \; k^\prime = 1, \ldots, N_t \\
\mathcal{R}_{t|t-1} (k) = \underset{k^\prime}{\mathrm{min}} \left(T_{k^\prime, k}^{(t|t-1)}\right)\; , \; k = 1, \ldots, M_{t|t-1}
\end{array}
\end{equation}

\noindent where $\mathcal{C}_{t|t-1} = \left[\mathcal{C}_{t|t-1} (1), \mathcal{C}_{t|t-1} (2), \ldots, \mathcal{C}_{t|t-1} (N_t)\right]^\intercal$ and \\ \noindent $\mathcal{R}_{t|t-1} = \left[\mathcal{R}_{t|t-1} (1), \mathcal{R}_{t|t-1} (2), \ldots, \mathcal{R}_{t|t-1} (M_{t|t-1}) \right]$ are $N_k \times 1$ and $1 \times M_{k-1}$ vectors, respectively. Each element of $\mathcal{C}_{t|t-1}$ and $\mathcal{R}_{t|t-1}$ quantifies the measurement-to-target and target-to-measurement closest distance, respectively.
In other words, $\mathcal{C}_{t|t-1} (k^\prime)$ and $\mathcal{R}_{t|t-1} (k)$ are the measurement and target errors for the ${k^\prime}^{th}$ measurement and $k^{th}$ target, respectively, which quantitatively indicate how genuine the found associated sample is.

Next, the histogram of $\mathcal{C}^{I}_{t|t-1}$ and $\mathcal{R}^{I}_{t|t-1}$ are computed as $\mathcal{H}^{\mathcal{C}}_{t|t-1}$ and $\mathcal{H}^{\mathcal{R}}_{t|t-1}$, respectively, as follows,
\begin{equation}
\begin{array}{l}
\mathcal{H}^{\mathcal{C}}_{t|t-1} = \mathrm{hist}\left({\mathcal{C}^{I}_{k|k-1}}^\intercal, {1:M_{t|t-1}}\right) \in \mathbb{Z}^{1 \times M_{t|t-1}}, \\
\mathcal{H}^{\mathcal{R}}_{t|t-1} = \mathrm{hist}\left(\mathcal{R}^{I}_{k|k-1}, {1:N_{t}}\right)  \in \mathbb{Z}^{1 \times N_{t}}.
\end{array}
\end{equation}
\noindent $\mathrm{hist}(\bullet, 1:M_{t|t-1})$ and $\mathrm{hist}(\bullet, 1:N_{t})$ compute the histogram of the input vectors by filling $[1, 2, \ldots, M_{t|t-1}]$ and $[1, 2, \ldots, N_{t}]$ bins, respectively.
The $k^{th}$ element of $\mathcal{H}^{\mathcal{C}}_{t|t-1}$ (i.e.\ $\mathcal{H}^{\mathcal{C}}_{t|t-1}(k)$) shows the number of associations for the $k^{th}$ predicted target. On the other hand, $\mathcal{H}^{\mathcal{R}}_{t|t-1}(k^\prime)$ indicates the number of association to the ${k^\prime}^{th}$ measurement. 

\noindent\textbf{\textit{Decay, survival and birth of targets.}}
The target tuples are then updated using the data association approach explained as a pseudo code in Algorithm~\ref{Al:dataAssociation}.
Basically, one of the following three hypotheses (cases in Algorithm~\ref{Al:dataAssociation}) are assigned for each filtered target: {\bf{Case (1)}} decaying status, which indicate the target has no association; {\bf{Case (2)}} survival status, for the targets with at least one measurement association; {\bf{Case (3)}} birth status, for those (isolated) measurements without any association.

\begin{algorithm}[!t]
\KwIn{$a_{min}, g_{min}, g_{max}, \mathcal{H}^{\mathcal{C}}_{t|t-1}, \mathcal{H}^{\mathcal{R}}_{t|t-1}, \mathcal{C}_{t|t-1},$ $\mathcal{R}_{t|t-1}, X_{t|t-1}, Z_t$}
\KwOut{Survived targets and births: $X_t^u$ and $X_t^b$}
\% Initialise the output RFS as empty sets of target tuples:\\
$X_t^u = \{()\}$;
$X_t^b = \{()\}$\\
\% Iterate over $M_{t|t-1}$ targets in $X_{t|t-1}$:\\
\For{$k = 1, 2, \ldots, M_{t|t-1}$}{
\If{($\mathcal{H}^{\mathcal{C}}_{t|t-1}(k) == 0$ AND $a_k^{(t-1)} \geq a_{min}$) OR ($\mathcal{H}^{\mathcal{C}}_{t|t-1}(k) \geq 1$ AND $g_{min} \leq \mathcal{R}_{t|t-1}(k) \leq g_{max}$) {\bf{(Case (1))}}}{
$\quad
\Bigg({\bf{m}}_k^{(t)} = \textrm{append}\Bigg({\bf{m}}_k^{(t-1)}, 
{\bf{\hat m}}_k^{(t|t-1)}\Bigg), 
a_k^{(t)} = a_k^{(t-1)} - 1,$\\ 
$\qquad \qquad g_k^{(t)} = \mathcal{R}_{t|t-1}(k), 
f_k^{(t)} = 1\Bigg)
$
$\rightarrow$ Append to $X_t^u$
}
\If{$\mathcal{H}^{\mathcal{C}}_{t|t-1}(k) \geq 1$ AND $\mathcal{R}_{t|t-1}(k) < g_{min}$ {\bf{(Case (2))}}}{
$\quad
\Bigg({\bf{m}}_k^{(t)} = \textrm{append}\Bigg({\bf{m}}_k^{(t-1)}, 
{\bf{z}}^{(t)}_{\mathcal{R}^{I}_{t|t-1} (k)}
\Bigg),$ \\
$\qquad a_k^{(t)} = a_k^{(t-1)} + 1, 
g_k^{(t)} = \mathcal{R}_{t|t-1}(k), 
f_k^{(t)} = 0\Bigg)
$
$\rightarrow$ Append to $X_t^u$
}
}
\% Iterate over $N_{t}$ measurements in $Z_t$:\\
\For{$k^\prime = 1, 2, \ldots, N_{t}$}{
\If{$\mathcal{H}^{\mathcal{R}}_{t|t-1}(k^\prime) == 0$ OR $\mathcal{C}_{t|t-1} (k') > g_{max}$ {\bf{(Case (3))}}}{
$\Bigg({\bf{m}}_{k^\prime}^{(t)} = {\bf{z}}^{(t)}_{k^\prime}, 
a_k^{(t)} = a_{min}, 
g_k^{(t)} = g_{min}, f_k^{(t)} = 0\Bigg)$
$\rightarrow$ App. to $X_t^b$
}
}
\caption{The data association algorithm.}
\label{Al:dataAssociation}
\end{algorithm}

\begin{figure}[!t]
\centering
\includegraphics[width=0.5\textwidth]{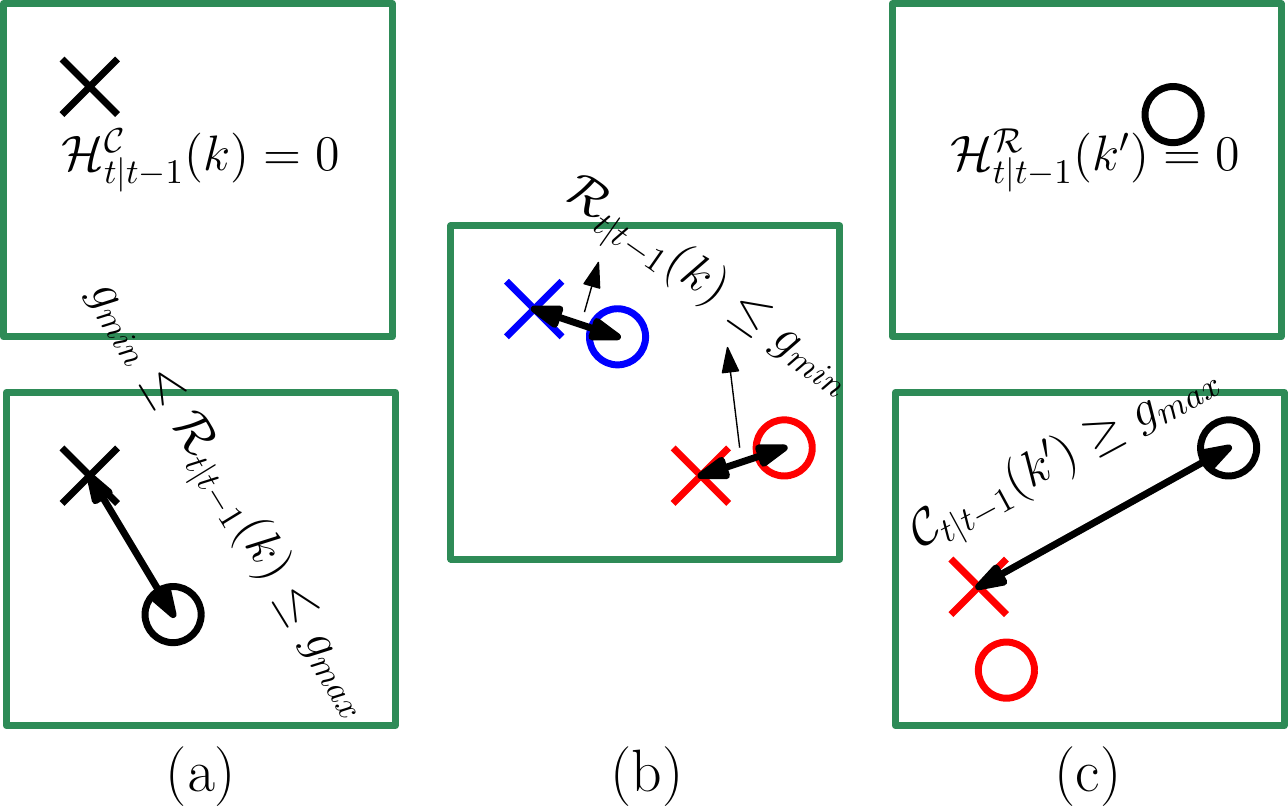}
\caption{Targets and measurements are shown as crosses and circles, respectively: (a) Decaying target: (a-top) target with no measurement; (a-bottom) target with a \emph{far} associated measurement; (b) Targets survival; (c) Birth of targets: (c-top) measurement with no associated target; (c-bottom) measurement with a distant associated target.}
\label{fig:DATAASSo}
\end{figure}

For {\bf{Case (1)}} shown in Fig.~\ref{fig:DATAASSo}-a, the freeze state is set to one, meaning that the association step failed to find a measurement sample for the current target (Fig.~\ref{fig:DATAASSo}-a-top) possibly due to occlusion, measurement failure or due to the fact that the target itself is a false positive (Fig.~\ref{fig:DATAASSo}-a-bottom where the associated measurement is far from the target $g_{min} \leq \mathcal{R}_{t|t-1}(k) \leq g_{max}$). In the absence of an associated measurement sample, the predicted target state ${\hat{\bf{m}}_k}^{(t|t-1)}$ is appended to ${\bf{m}}_k^{(t-1)}$ to create the new state matrix: 
${\bf{m}}_k^{(t)} = \left[\begin{smallmatrix}
{\bf{m}}_k^{(t-1)}\\
----\\
{\hat{\bf{m}}_k}^{(t|t-1)}
\end{smallmatrix}\right].$
For {\bf{Case (2)}} illustrated in Fig.~\ref{fig:DATAASSo}-b, the freeze state of the target is set to zero as the target is associated with at least one measurement ($\mathcal{R}_{t|t-1}(k) < g_{min}$). Its target state matrix ${\bf{m}}_k^{(t-1)}$ is updated by appending the associated measurement vector ${\bf{z}}_{k^\prime}^{(t)}$, i.e. 
${\bf{m}}_k^{(t)} = \left[\begin{smallmatrix}
{\bf{m}}_k^{(t-1)}\\
----\\
{\bf{z}}_{k^\prime}^{(t)}
\end{smallmatrix}\right].$  
For both cases, to optimise memory allocation we define a maximum batch size.
If the number of rows in ${\bf{m}}_k^{(t)}$ (i.e.\ $S_k^{(t)}$) was greater than a maximum assigned batch size, the first row of ${\bf{m}}_k^{(t)}$ which corresponds to the oldest saved prediction or measurement is removed.
The assigned target tuples form the updated RFS $X^{u}_t$ as explained in Algorithm~\ref{Al:dataAssociation}.

In parallel with the above two procedures, the third case (Fig.~\ref{fig:DATAASSo}-c) is evaluated to determine birth of targets. For a measurement with no target association ($\mathcal{H}^{\mathcal{R}}_{t|t-1}(k^\prime) = 0$ shown at Fig.~\ref{fig:DATAASSo}-c-top) or an isolated measurement (whose $\mathcal{C}_{t|t-1}(k^\prime) > g_{max}$, shown at Fig.~\ref{fig:DATAASSo}-c-bottom), a new target tuple is assigned. Concatenating all of these tuples form the target birth RFS $X^b_{t}$.
The target tuple at the $t^{th}$ time step is calculated as the union of births and survivals, i.e. $X_t = X^b_{t} \cup X^u_{t}$, which has $M_t$ cardinality.

\noindent {\bf{\textit{On the data association algorithm complexity.}}}
The complexity of similar assignment method, such as the Hungarian matching derivations, can reach to $\mathcal{O}(n^4)$ \cite{Liu:2010}. Also, assuming no measurement gating is performed, the computational complexity of GM-PHD filter is $\approx \mathcal{O}(M_{t|t-1} \times N_{t})$ \cite{Vo:2006}. On the other hand, the proposed data association has $\mathcal{O}\left(M_{t|t-1} + N_{t}\right)$ complexity, while implementing the two loops in Algorithm \ref{Al:dataAssociation} in parallel can reduce the complexity to $\mathcal{O}\left({\mathrm{max}}\left([M_{t|t-1}, N_{t}]\right) \right)$.

\section{Experimental results}
\label{sec:expRes}
\subsection{Datasets and evaluation metric}
The proposed MTF algorithm is evaluated over two data sequences: (1) a controlled simulation MTF introduced by Vo and Ma~\cite{Reuter:2014,Vo:2014}; (2) A bird's-eye view representation of the targets in the DukeMTMC dataset. 
We compute the Optimal Sub-Pattern Assignment (OSPA, \cite{Schuhmacher:2008}) distance to quantitatively evaluate the proposed algorithm. The OSPA error consists of two terms: one is related to the difference in the number compared sets (cardinality (Card) error); and the other relates to the localisation $cost$ (Loc), which is the smallest pair-wise distance among all the elements in the two sets. In our work, we have used the Hungarian assignment to compute this minimal distance. OSPA has been widely used for evaluating the accuracy of the point target filtering  algorithms~\cite{Vo:2017,Fantacci:2018}. 
The overall pipeline is implemented (end-to-end) in Python 2.7, and all the experiments are tested using an NVIDIA GeForce GTX 1080 GPU and an $i5-8400$ CPU. We have used a 3-layer LSTM network ($L=3$), each having 20 hidden units, outputting a fully-connected layer (\ref{eq:LSTM_out}), with $\phi^{(y)}$ as an identity function. 
The network is trained online over the currently updated patch for each target, minimising the mean square error as the loss function and using Adam optimisation method. The training procedure is performed with 0.001 learning rate, and optimiser parameters of $\beta_1=0.9$ and $\beta_2=0.99$. As in \cite{Schuhmacher:2008}, we choose $p=1$ and $c=100$.

The process of initialising the LSTM by allocating GPU memory via Tensorflow, training it as a regression block and predicting the output sample takes $\approx 0.3$ millisecond for each target, while it takes $\approx 0.1$ milliseconds (per target) to fine-tune the LSTM network. 
To be more specific, as described in Section~\ref{sec:LSTMTrain}, at every time step, we transfer the weights and biases ($\mathcal{M}_{t|t-1}$) learned from the motion trajectories of other targets to the current one and only fine-tune its weights and biases using fewer number of epochs. This reduces the computation time from $\approx 0.3$ milliseconds (elapsed time to initialise the LSTM, allocate the GPU memory, train as regression block using 50 epochs and predict the output sample) to $\approx 0.1$ millisecond for each target, with 20 epochs (Please see our supplementary material for video samples).

\subsection{Results on synthetic data}
\begin{figure}[!b]
\centering
\includegraphics[width=0.5\textwidth]{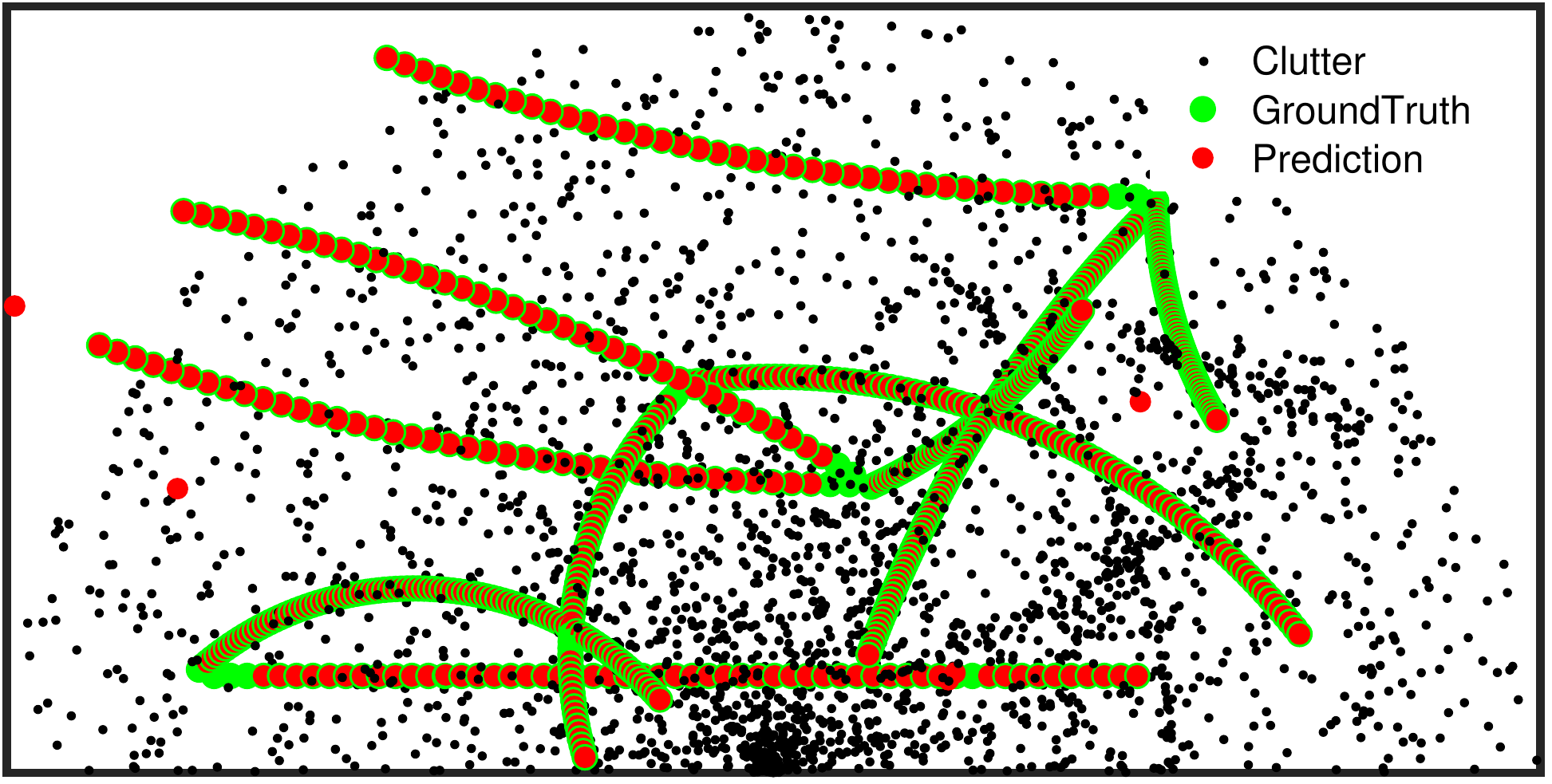}
\caption{Temporally overlaid visualisation of the target trajectories: The ground truth, cluttered noisy measurements and OLMM filtering results are shown in green, black and red, respectively.}
\label{fig:SimEnvRes}
\end{figure}

In this scenario, there are 10 targets appearing in the scene at different times, having various birth times and lifespans (Fig.~\ref{fig:SimEnvRes}). 
The measurements is performed by computing the range and bearing (azimuth) of a target from the origin. It also contains clutter with uniform distribution along range and azimuth, with a random intensity sampled from a Poisson distribution with $\lambda_c$ mean. The obtained measurements are degraded by a Gaussian noise with zero mean and $\sigma_r=10$ (unit distance) and $\sigma_\theta=\pi/90$ (rad) standard deviation, respectively. The problem is to perform online MTF to recover true positives from clutter.
In our first experiment we compute the OSPA error, assuming $\lambda_c = 20$ clutter intensity. 

\begin{figure}[!t]
\centering
\includegraphics[width=0.5\textwidth]{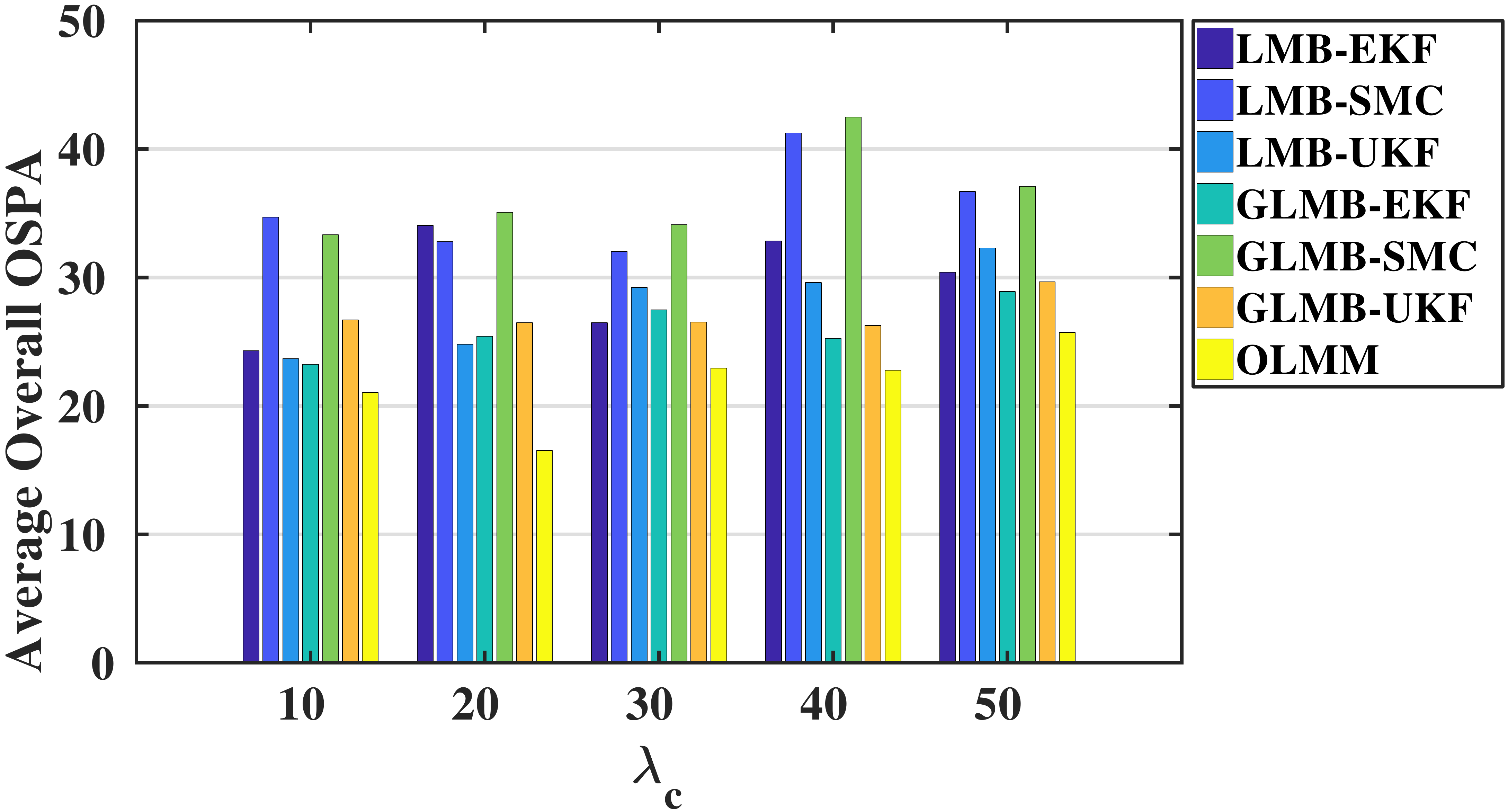}
\caption{Average overall OSPA for different methods for different $\lambda_c$.}
\label{fig:OSPAClutter}
\end{figure}

\begin{table}[!t]
\begin{center}
\begin{tabular}{c | c c c} 
Algorithm & OSPA Card & OSPA Loc & OSPA \\
\hline
PHD-EKF & $9.25$ & $20.86$ & $30.11$ \\
PHD-SMC & $12.76$ & $46.08$ & $58.84$ \\
PHD-UKF & $10.33$ & $19.73$ & $30.06$ \\
\hdashline
CPHD-EKF & $7.10$ & $23.00$ & $30.10$ \\
CPHD-SMC & $11.18$ & $46.08$ & $57.25$ \\
CPHD-UKF & $5.50$ & $22.39$ & $27.89$ \\
\hdashline
LMB-EKF & $4.59$ & $22.59$ & $27.18$ \\
LMB-SMC & $12.07$ & $23.47$ & $35.54$ \\
LMB-UKF & $3.77$ & $21.94$ & $25.72$ \\
\hdashline
GLMB-EKF & $6.37$ & $20.13$ & $26.50$ \\
GLMB-SMC & $6.11$ & $21.07$ & $27.19$ \\
GLMB-UKF & $11.79$ & $19.84$ & $31.63$ \\
\hdashline
\textbf{OLMM} & $7.33$ & $9.93$ & $\textbf{17.26}$
\end{tabular}
\end{center}
\caption{OSPA error for different methods over the synthetic multi-target scenario: we compared our approach to PHD, CPHD \cite{Nagappa:2017,Mahler:2007}, LMB \cite{Reuter:2014}, and GLMB \cite{Vo:2014,Vo:2017} algorithm, when EKF, SMC, and UKF are used for prediction and update steps.}
\label{tab:table1}
\end{table}

\begin{figure}[!t]
    \centering
    \begin{subfigure}[b]{0.5\textwidth}
        \includegraphics[width=\textwidth]{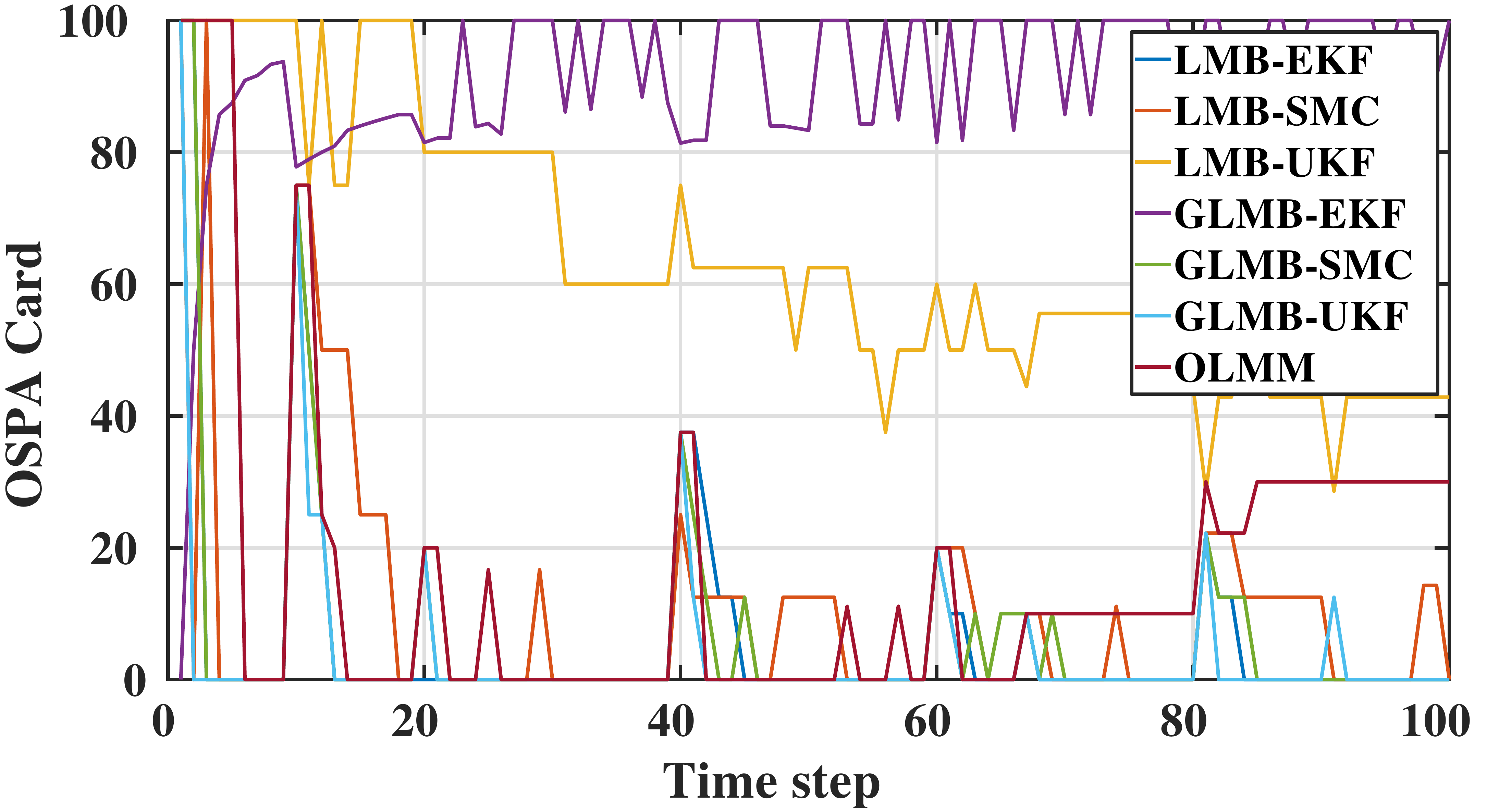}
        \caption{}
    \end{subfigure}
    \begin{subfigure}[b]{0.5\textwidth}
        \includegraphics[width=\textwidth]{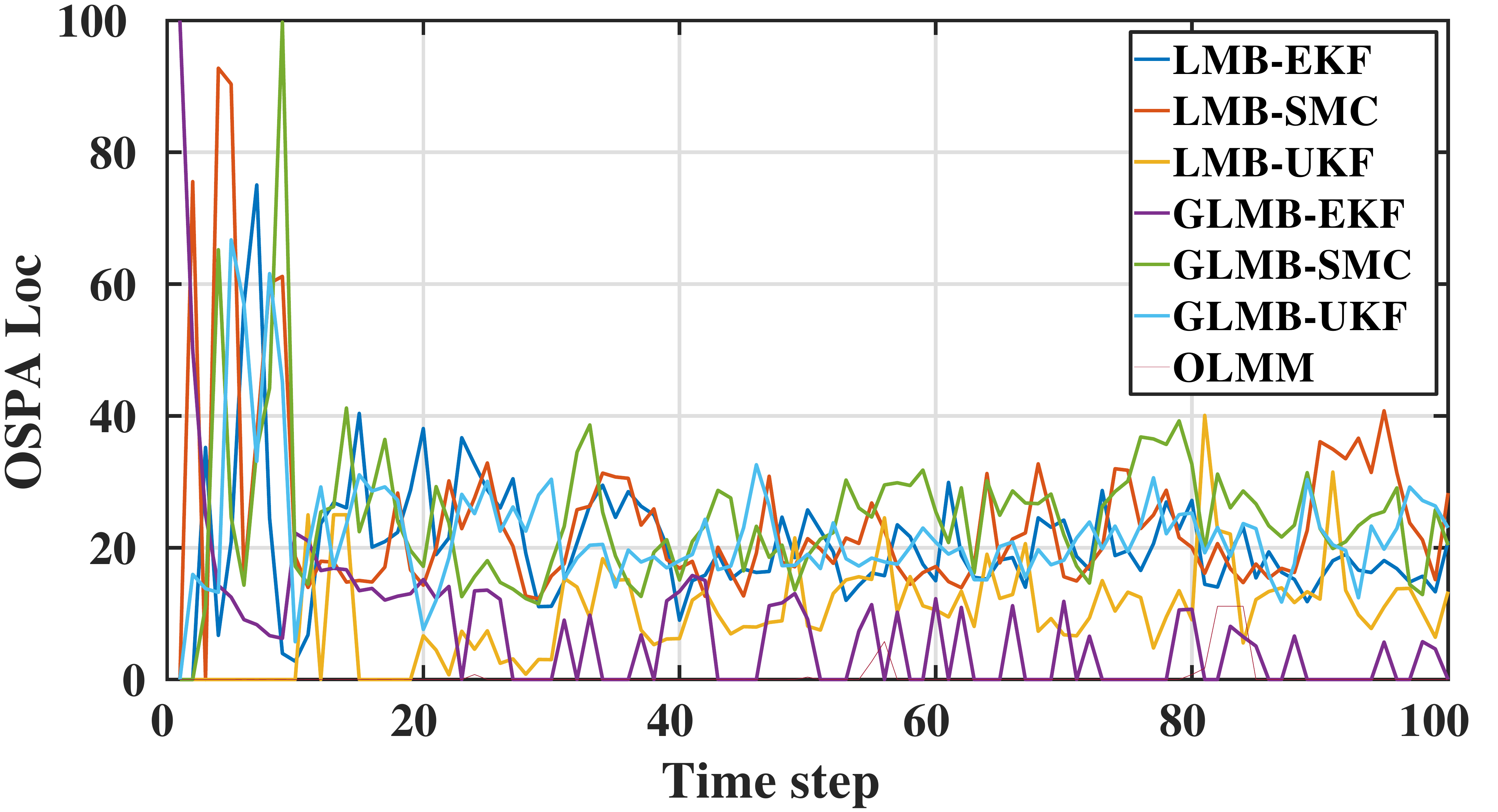}
        \caption{}
    \end{subfigure}
    \begin{subfigure}[b]{0.5\textwidth}
        \includegraphics[width=\textwidth]{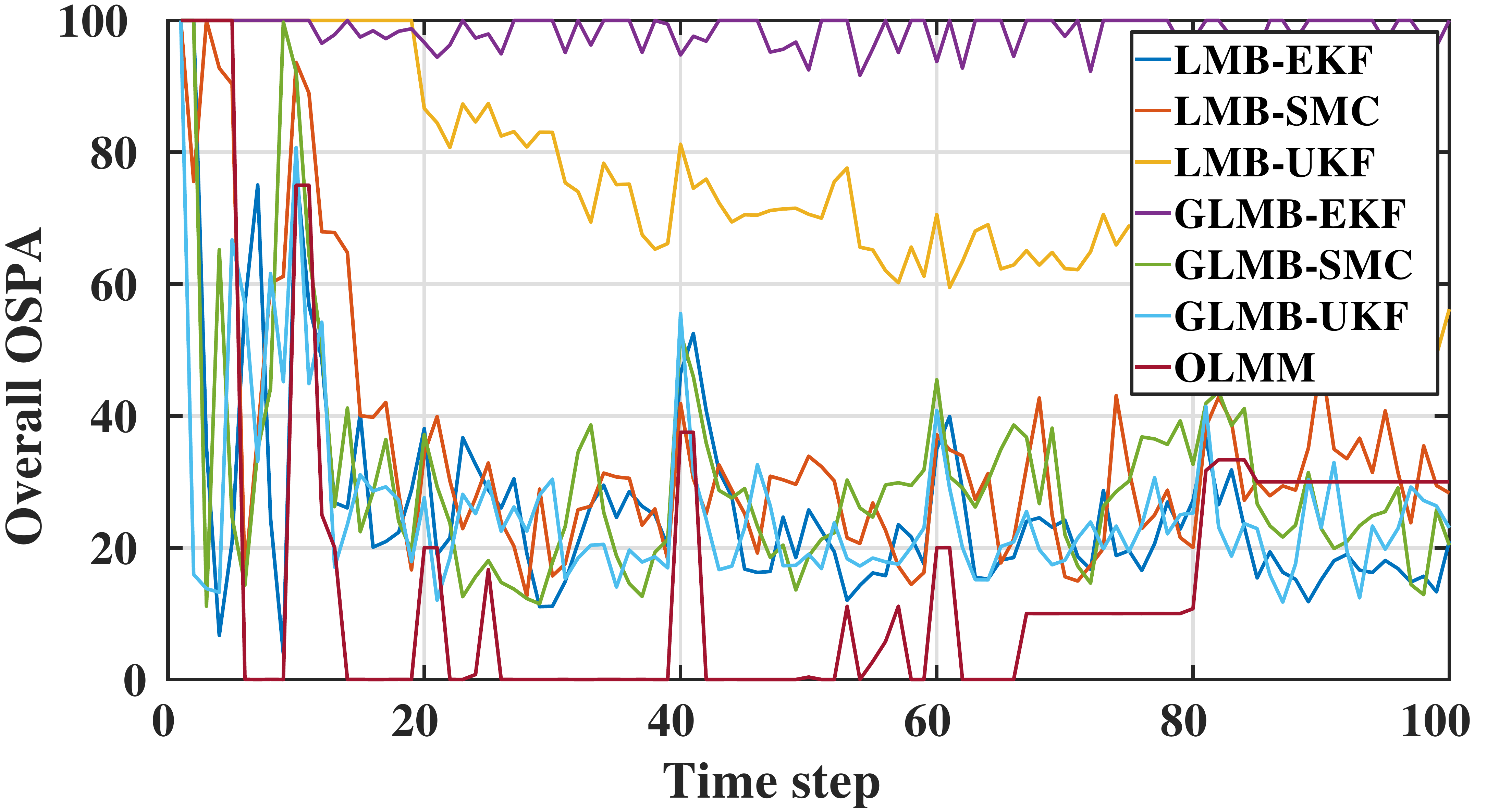}
        \caption{}
    \end{subfigure}
    \caption{Comparison of overall OSPA error for different methods for $\lambda_c=20$: (a) Card; (b) Loc and (c) Overall OSPA errors.}
    \label{fig:OSPA_time}
\end{figure}

In Table \ref{tab:table1}, we report the average overall OSPA and its two terms related to cardinality error (OSPA Card) and optimal Hungarian distance (OSPA Loc - the $cost$ term). We compare our method with PHD, CPHD, LMB, and GLMB algorithm, when EKF, SMC, and UKF used as basis for the prediction and update steps (The following Matlab implementation of these algorithms is used: \url{http://ba-tuong.vo-au.com/codes.html}). Our method outperforms all the other algorithms in terms of overall OSPA. In particular, this is due to a significant drop of the Loc error, while cardinality error is comparable with most of the others. 

\begin{figure}[!t]
\centering
\includegraphics[width=0.5\textwidth]{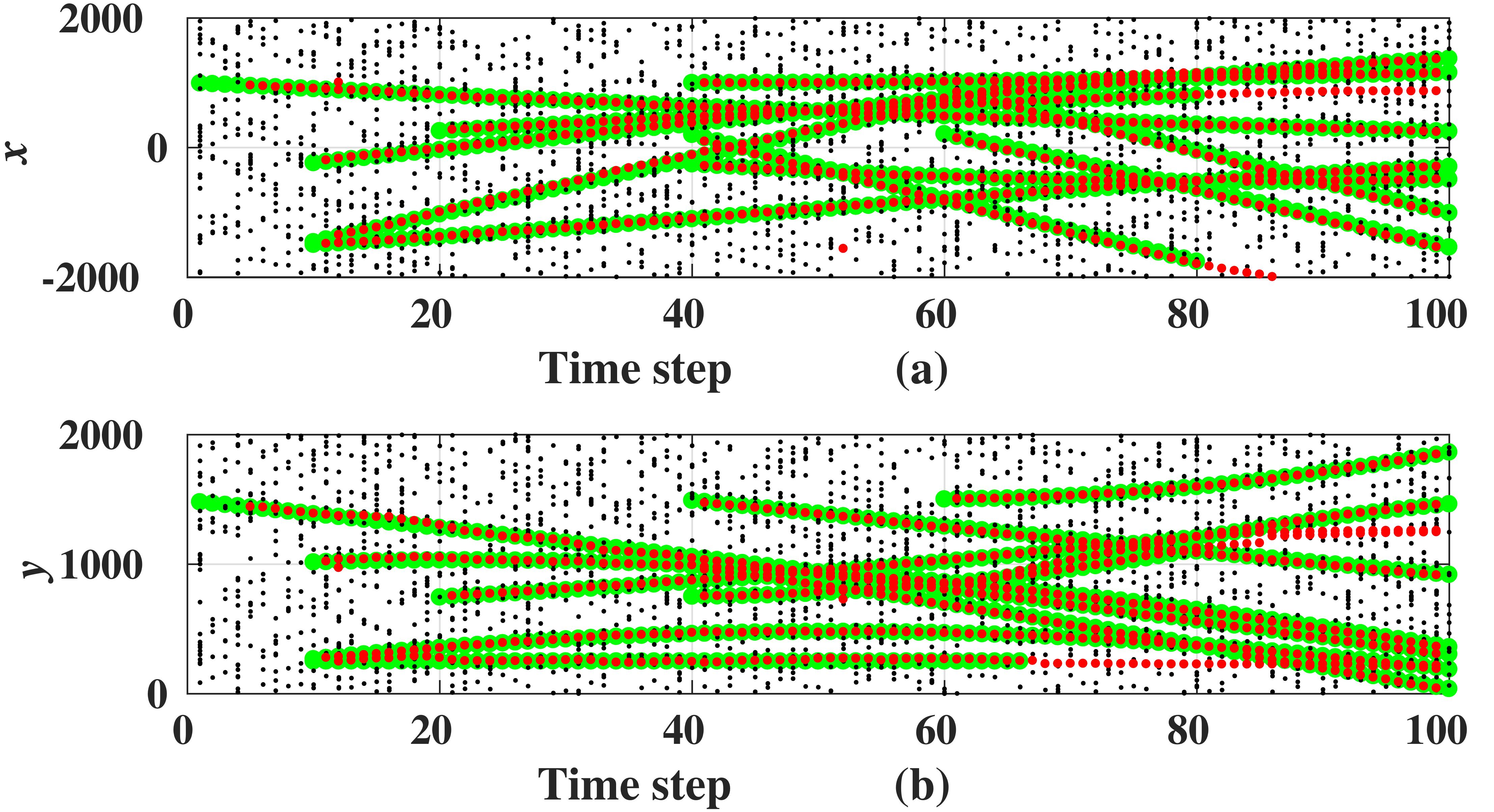}
\caption{Projection of clutter (black), targets (red) and ground truth (green) points onto the: (a) $x-$ and (b) $y-$ axes.}
\label{fig:XYProj}
\end{figure}

The resulting trajectories of this experiment for our method are illustrated in Fig.~\ref{fig:SimEnvRes}. The red dots represent the predicted location of the targets at every time step, filtered out from the measurements clutters (black dots). They almost overlap with the ground truth (green dots), except very few (only three) false positives (predicted but no ground truth) and false negatives (ground truth but no prediction). The target and clutter data projections onto the horizontal and vertical axes at each time step are also plotted in Fig.~\ref{fig:XYProj}.

Moreover, in Fig.~\ref{fig:OSPA_time} we show the overall OSPA at every time steps. During the initial time steps (frame number $ < 6$), our OSPA error is higher. This is mostly due to the under-fitting of the LSTM model because of lack of data. However, after $\approx 7^{th}$ time step our OSPA error becomes significantly lower than other approaches, having an overall average of $\approx 18$, while the average OSPA error for other algorithms are $> 25$. 
The impulsive peaks correspond to those time steps when birth of targets occur, at which the Card error suddenly increases.
In order to show the robustness of our algorithm for higher clutter densities, in the second experiment, we increase the clutter intensity $\lambda_c$ and find the average OSPA over all time steps. Figure~\ref{fig:OSPAClutter} shows the results of this experiment, for $\lambda_c = 10, 20, \ldots, 50$. Our filtering algorithm provides a relatively constant and comparably lower overall OSPA error even when the clutter is increased to 50. Both of the SMC-based algorithms (GLMB-SMC and LMB-SMC) generate highest OSPA error, which can be due to the particle filter algorithm divergence. On the other hand, lower OSPA errors generated by the LMB with an EKF model shows how successfully this particular simulated scenario can be modelled using such non-linear filter. 
It should be mentioned, however, that our method does not rely on any prior motion model capable of learning the non-linearity within the data sequence.

\subsection{Results on the Duke dataset}
DukeMTMC is a pedestrian tracking dataset, captured using 8 synchronised cameras \cite{Ristani:2016}. 
In our experiments, we use its 177840 frames, whose ground truth are provided. In order to evaluated our point target filtering algorithm, we map the coordinates of the bottom centre of each bounding box to an aerial perspective. Each of these points are first mapped from the image plane to the world coordinate system, and then to the aerial map (the bird's-eye view map).
We repeat the same procedure over the provided OpenPose \cite{Cao:2017} detection results (used as measurement sets in our experiments) to obtain their corresponding aerial view representation.

\begin{figure*}[!t]
\centering
\includegraphics[width=1.0\textwidth]{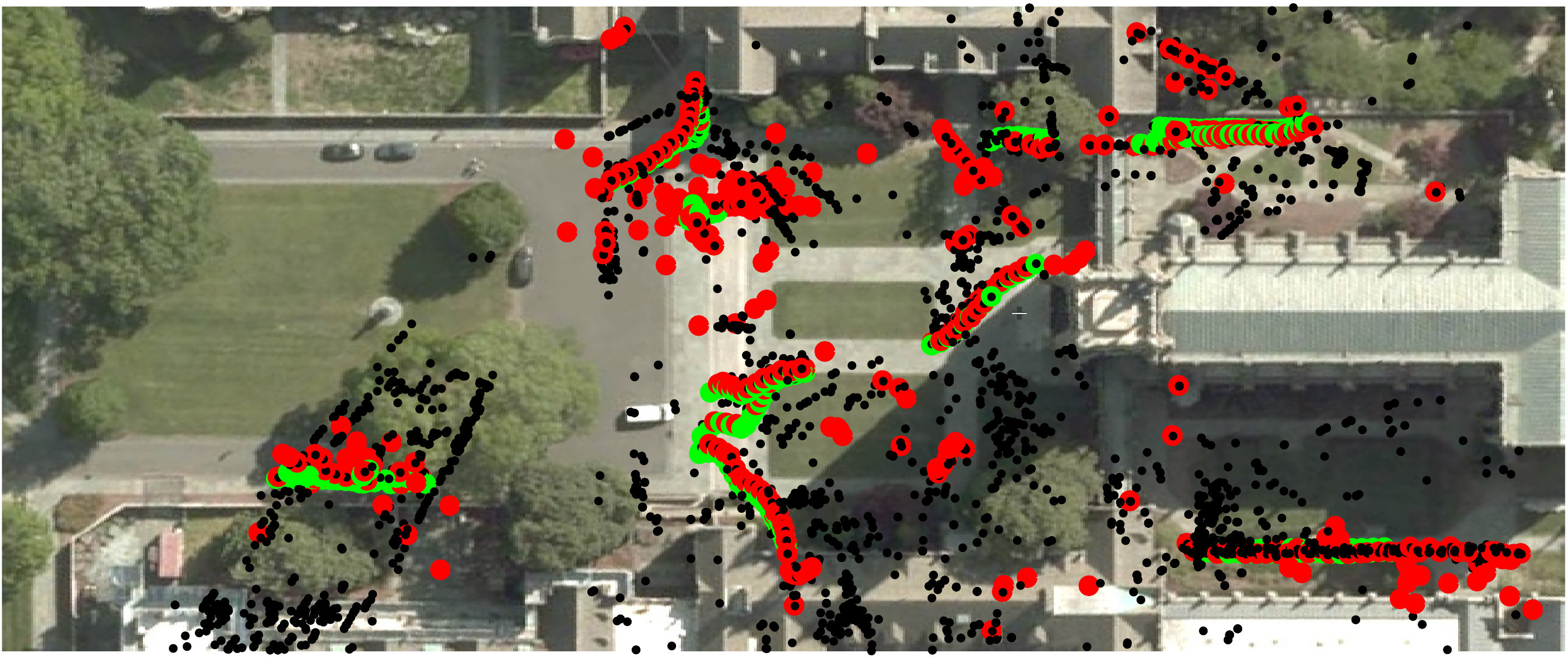}
\caption{Bird's-eye view of point targets in DukeMTMC: The green, black and red points are the ground truth, detection and OLMM results, respectively.}
\label{fig:DukeOverlay}
\end{figure*}

The computed trajectories of this experiment for our method are shown in Fig.~\ref{fig:DukeOverlay}. The red dots represent the target locations given by OLMM at every time step, filtering the highly cluttered measurements (black dots). The filtered point targets almost totally overlap with the ground truth (green dots), except for some false positive targets. As the camera locations are fixed, such targets are mainly caused by persistent false detections.

The MTF performance of several algorithms are quantitatively illustrated in Table~\ref{tab:table2} in term of their OSPA errors. For each algorithm, the OSPA errors are calculated for each data frame and then averaged for the whole 177840 sequence.
The performance of OLMM outperforms all the other algorithms (PHD, CPHD, LMB, and GLMB, when EKF, SMC, and UKF are used as basis for the prediction and update steps). OLMM generates the third lowest cardinality error, while simultaneously maintaining a low OSPA Loc error, resulting in the lowest overall OSPA. 

\begin{table}[!t]
\begin{center}
\begin{tabular}{c | c c c} 
Algorithm & OSPA Card & OSPA Loc & OSPA \\
\hline
PHD-EKF & $38.33$ & $25.30$ & $63.63$ \\
PHD-SMC & $63.90$ & $7.41$ & $71.30$ \\
PHD-UKF & $59.21$ & $5.22$ & $64.42$ \\
\hdashline
CPHD-EKF & $46.16$ & $13.37$ & $59.53$ \\
CPHD-SMC & $46.16$ & $20.43$ & $66.59$ \\
CPHD-UKF & $46.16$ & $13.58$ & $59.74$ \\
\hdashline
LMB-EKF & $72.78$ & $17.22$ & $90.00$ \\
LMB-SMC & $32.34$ & $55.55$ & $87.90$ \\
LMB-UKF & $74.08$ & $18.04$ & $92.13$ \\
\hdashline
GLMB-EKF & $58.61$ & $16.25$ & $74.85$ \\
GLMB-SMC & $80.95$ & $7.58$ & $88.53$ \\
GLMB-UKF & $81.91$ & $6.74$ & $88.65$ \\
\hdashline
\textbf{OLMM} & $41.76$ & $14.84$ &  $\textbf{56.61}$
\end{tabular}
\end{center}
\caption{OSPA error for different methods over the DukeMTMC dataset.}
\label{tab:table2}
\end{table}

\section{Conclusions}
\label{sec:Conc}
This paper proposes an MTF algorithm which learns the motion models, on the fly, using an RNN with an LSTM architecture, as a regression problem. The target state predictions are then corrected using a novel data association algorithm, with a low computational complexity. The proposed algorithm is evaluated over synthetic and real point target filtering scenarios, demonstrating a remarkable performance over highly cluttered data sequences.

The proposed OLMM algorithm can be applied to various applications where point targets are obtained by the detectors. Some examples can be target tracking from satellite images, keypoint filtering for {3D} scene mapping, radar point scatterer detection and tracking, and LiDAR signal processing. Also, as the proposed approach does not assign limits over the state space dimensionality, in addition to point target filtering, the algorithm's potential to filter extended targets can be investigated.

\bibliographystyle{IEEEtran}
\bibliography{refs}

\begin{thebibliography}{10}
\providecommand{\url}[1]{#1}
\csname url@samestyle\endcsname
\providecommand{\newblock}{\relax}
\providecommand{\bibinfo}[2]{#2}
\providecommand{\BIBentrySTDinterwordspacing}{\spaceskip=0pt\relax}
\providecommand{\BIBentryALTinterwordstretchfactor}{4}
\providecommand{\BIBentryALTinterwordspacing}{\spaceskip=\fontdimen2\font plus
\BIBentryALTinterwordstretchfactor\fontdimen3\font minus
  \fontdimen4\font\relax}
\providecommand{\BIBforeignlanguage}[2]{{%
\expandafter\ifx\csname l@#1\endcsname\relax
\typeout{** WARNING: IEEEtran.bst: No hyphenation pattern has been}%
\typeout{** loaded for the language `#1'. Using the pattern for}%
\typeout{** the default language instead.}%
\else
\language=\csname l@#1\endcsname
\fi
#2}}
\providecommand{\BIBdecl}{\relax}
\BIBdecl

\bibitem{Li:2018}
S.~{Li}, W.~{Yi}, R.~{Hoseinnezhad}, B.~{Wang}, and L.~{Kong}, ``Multiobject
  tracking for generic observation model using labeled random finite sets,''
  \emph{IEEE Transactions on Signal Processing}, vol.~66, no.~2, pp. 368--383,
  2018.

\bibitem{Punchihewa:2018}
Y.~G. {Punchihewa}, B.~{Vo}, B.~{Vo}, and D.~Y. {Kim}, ``Multiple object
  tracking in unknown backgrounds with labeled random finite sets,'' \emph{IEEE
  Transactions on Signal Processing}, vol.~66, no.~11, pp. 3040--3055, 2018.

\bibitem{Roy:2016}
A.~Roy and D.~Mitra, ``Multi-target trackers using cubature kalman filter for
  doppler radar tracking in clutter,'' \emph{IET Signal Processing}, vol.~10,
  pp. 888--901(13), 2016.

\bibitem{Kulikov:2016}
G.~Y. {Kulikov} and M.~V. {Kulikova}, ``The accurate continuous-discrete
  extended {K}alman filter for radar tracking,'' \emph{IEEE Transactions on
  Signal Processing}, vol.~64, no.~4, pp. 948--958, 2016.

\bibitem{Evers:2018}
C.~{Evers} and P.~A. {Naylor}, ``Optimized self-localization for {SLAM} in
  dynamic scenes using probability hypothesis density filters,'' \emph{IEEE
  Transactions on Signal Processing}, vol.~66, no.~4, pp. 863--878, 2018.

\bibitem{Leung:2017}
K.~Y.~K. {Leung}, F.~{Inostroza}, and M.~{Adams}, ``Relating random vector and
  random finite set estimation in navigation, mapping, and tracking,''
  \emph{IEEE Transactions on Signal Processing}, vol.~65, no.~17, pp.
  4609--4623, 2017.

\bibitem{Fantacci:2018}
C.~Fantacci, B.~N. Vo, B.~T. Vo, G.~Battistelli, and L.~Chisci, ``Robust fusion
  for multisensor multiobject tracking,'' \emph{IEEE Signal Processing
  Letters}, vol.~25, no.~5, pp. 640--644, 2018.

\bibitem{Li:20182}
S.~{Li}, W.~{Yi}, R.~{Hoseinnezhad}, G.~{Battistelli}, B.~{Wang}, and
  L.~{Kong}, ``Robust distributed fusion with labeled random finite sets,''
  \emph{IEEE Transactions on Signal Processing}, vol.~66, no.~2, pp. 278--293,
  2018.

\bibitem{Xing:2016}
Z.~Xing and Y.~Xia, ``Comparison of centralised scaled unscented {K}alman
  filter and extended {K}alman filter for multisensor data fusion
  architectures,'' \emph{IET Signal Processing}, vol.~10, pp. 359--365(6),
  2016.

\bibitem{Yan:2018}
L.~Yan, L.~Jiang, J.~Liu, Y.~Xia, and M.~Fu, ``Optimal distributed {K}alman
  filtering fusion for multirate multisensor dynamic systems with correlated
  noise and unreliable measurements,'' \emph{IET Signal Processing}, vol.~12,
  pp. 522--531(9), 2018.

\bibitem{Roth:2014}
M.~Roth, G.~Hendeby, and F.~Gustafsson, ``{EKF}/{UKF} maneuvering target
  tracking using coordinated turn models with polar/{C}artesian velocity,'' in
  \emph{17th International Conference on Information Fusion (FUSION)}.\hskip
  1em plus 0.5em minus 0.4em\relax IEEE, 2014, pp. 1--8.

\bibitem{Li:2000}
X.~R. Li and V.~P. Jilkov, ``Survey of maneuvering target tracking: dynamic
  models,'' in \emph{Signal and Data Processing of Small Targets 2000}, vol.
  4048.\hskip 1em plus 0.5em minus 0.4em\relax International Society for Optics
  and Photonics, 2000, pp. 212--236.

\bibitem{Zhai:2014}
G.~Zhai, H.~Meng, and X.~Wang, ``A constant speed changing rate and constant
  turn rate model for maneuvering target tracking,'' \emph{Sensors}, vol.~14,
  no.~3, pp. 5239--5253, 2014.

\bibitem{Reuter:2014}
S.~Reuter, B.~T. Vo, B.~N. Vo, and K.~Dietmayer, ``The labeled
  multi-{B}ernoulli filter,'' \emph{IEEE Transactions on Signal Processing},
  vol.~62, no.~12, pp. 3246--3260, 2014.

\bibitem{Vo:2014}
B.~N. Vo, B.~T. Vo, and D.~Phung, ``Labeled random finite sets and the {Bayes}
  multi-target tracking filter,'' \emph{IEEE Transactions on Signal
  Processing}, vol.~62, no.~24, pp. 6554--6567, 2014.

\bibitem{Vo:2006}
B.~N. Vo and W.~K. Ma, ``The {G}aussian mixture probability hypothesis density
  filter,'' \emph{IEEE Transactions on Signal Processing}, vol.~54, no.~11, pp.
  4091--4104, 2006.

\bibitem{Milan:2017}
A.~Milan, S.~Rezatofighi, A.~Dick, I.~Reid, and K.~Schindler, ``Online
  multi-target tracking using recurrent neural networks,'' in \emph{AAAI},
  2017.

\bibitem{Emambakhsh:2019}
M.~Emambakhsh, A.~Bay, and E.~Vazquez, ``Deep recurrent neural network for
  multi-target filtering,'' in \emph{International Conference on MultiMedia
  Modeling (MMM)}, 2019, pp. 519--531.

\bibitem{Mahler:2003}
R.~P.~S. Mahler, ``Multitarget {Bayes} filtering via first-order multitarget
  moments,'' \emph{IEEE Transactions on Aerospace and Electronic Systems},
  vol.~39, no.~4, pp. 1152--1178, 2003.

\bibitem{Vo:2005}
B.~N. Vo, S.~Singh, and A.~Doucet, ``Sequential {M}onte {C}arlo methods for
  multitarget filtering with random finite sets,'' \emph{IEEE Transactions on
  Aerospace and Electronic Systems}, vol.~41, no.~4, pp. 1224--1245, 2005.

\bibitem{Mahler:2007}
R.~Mahler, ``{PHD} filters of higher order in target number,'' \emph{IEEE
  Transactions on Aerospace and Electronic Systems}, vol.~43, no.~4, pp.
  1523--1543, 2007.

\bibitem{Nagappa:2017}
S.~Nagappa, E.~D. Delande, D.~E. Clark, and J.~Houssineau, ``A tractable
  forward-backward {CPHD} smoother,'' \emph{IEEE Transactions on Aerospace and
  Electronic Systems}, vol.~53, no.~1, pp. 201--217, 2017.

\bibitem{Lu:2017}
Z.~Lu, W.~Hu, and T.~Kirubarajan, ``Labeled random finite sets with moment
  approximation,'' \emph{IEEE Transactions on Signal Processing}, vol.~65,
  no.~13, pp. 3384--3398, 2017.

\bibitem{Bryant:2017}
D.~S. {Bryant}, E.~D. {Delande}, S.~{Gehly}, J.~{Houssineau}, D.~E. {Clark},
  and B.~A. {Jones}, ``The {CPHD} filter with target spawning,'' \emph{IEEE
  Transactions on Signal Processing}, vol.~65, no.~5, pp. 13\,124--13\,138,
  2017.

\bibitem{Schlangen:2018}
I.~{Schlangen}, E.~D. {Delande}, J.~{Houssineau}, and D.~E. {Clark}, ``A
  second-order {PHD} filter with mean and variance in target number,''
  \emph{IEEE Transactions on Signal Processing}, vol.~66, no.~1, pp. 48--63,
  2018.

\bibitem{Garc:2018}
Ã.~F. {Garc\'{i}a-Fern\'{a}ndez}, J.~L. {Williams}, K.~{Granström}, and
  L.~{Svensson}, ``Poisson multi-{B}ernoulli mixture filter: Direct derivation
  and implementation,'' \emph{IEEE Transactions on Aerospace and Electronic
  Systems}, vol.~54, no.~4, pp. 1883--1901, 2018.

\bibitem{Parisini:1994}
T.~Parisini and R.~Zoppoli, ``Neural networks for nonlinear state estimation,''
  \emph{International Journal of Robust and Nonlinear Control}, vol.~4, no.~2,
  pp. 231--248, 1994.

\bibitem{Bay2016}
A.~Bay, S.~Lepsoy, and E.~Magli, ``Stable limit cycles in recurrent neural
  networks,'' in \emph{2016 International Conference on Communications (COMM)},
  2016, pp. 89--92.

\bibitem{Liu:2010}
L.~Liu and D.~Shell, ``Assessing optimal assignment under uncertainty: An
  interval-based algorithm,'' in \emph{Proceedings of Robotics: Science and
  Systems}, 2010.

\bibitem{Schuhmacher:2008}
D.~Schuhmacher, B.~T. Vo, and B.~N. Vo, ``A consistent metric for performance
  evaluation of multi-object filters,'' \emph{IEEE Transactions on Signal
  Processing}, vol.~56, no.~8, pp. 3447--3457, 2008.

\bibitem{Vo:2017}
B.~N. Vo, B.~T. Vo, and H.~G. Hoang, ``An efficient implementation of the
  generalized labeled multi-{B}ernoulli filter,'' \emph{IEEE Transactions on
  Signal Processing}, vol.~65, no.~8, pp. 1975--1987, 2017.

\bibitem{Ristani:2016}
E.~Ristani, F.~Solera, R.~Zou, R.~Cucchiara, and C.~Tomasi, ``Performance
  measures and a data set for multi-target, multi-camera tracking,'' in
  \emph{European Conference on Computer Vision workshop on Benchmarking
  Multi-Target Tracking}, 2016.

\bibitem{Cao:2017}
Z.~Cao, T.~Simon, S.-E. Wei, and Y.~Sheikh, ``Realtime multi-person {2D} pose
  estimation using part affinity fields,'' in \emph{CVPR}, 2017.

\end{thebibliography}




\end{document}